\documentclass[letterpaper]{article}
\usepackage{aaai17}
\usepackage{times}
\usepackage{helvet}
\usepackage{courier}
\usepackage{amsmath}
\usepackage{graphicx}
\usepackage{algorithmic}
\usepackage[ruled]{algorithm2e}
\usepackage{makecell}
\begin{document}
%
\title{Automatically Building Face Datasets of New Domains \\from Weakly Labeled Data with Pretrained Models}
\author{Shengyong Ding \and Junyu Wu \and Wei Xu \and Hongyang Chao \\
	Sun Yat-sen University
}
\maketitle
\begin{abstract}

Training data are critical in face recognition systems. However, labeling a large scale face data for a particular domain is very tedious. In this paper, we propose a method to automatically and incrementally construct datasets from massive weakly labeled data of the target domain which are readily available on the Internet under the help of a pretrained face model. More specifically, given a large scale weakly labeled dataset in which each face image is associated with a label, i.e. the name of an identity, we create a graph for each identity with edges linking matched faces verified by the existing model under a tight threshold. Then we use the maximal subgraph as the cleaned data for that identity. With the cleaned dataset, we update the existing face model and use the new model to filter the original dataset to get a larger cleaned dataset. We collect a large weakly labeled dataset containing 530,560 Asian face images of 7,962 identities from the Internet, which will be published for the study of face recognition. By running the filtering process, we obtain a cleaned datasets (99.7+\% purity) of size  223,767 (recall 70.9\%).
On our testing dataset of Asian faces, the model trained by the cleaned dataset achieves recognition rate 93.1\%, which obviously outperforms the model trained by the public dataset CASIA  whose recognition rate is 85.9\%.

\end{abstract}

\section{Introduction}
Face recognition, i.e. determining a pair of face images are from the same person is a
central task in a lot of vision based applications. Recently, dramatic progress has been made by applying deep
learning methods \cite{taigman2014deepface,sun2014deep,yi2014learning,DBLP:FaceNet} with millions of training data. While
these models are promising in experiments on the public testing datasets, they still suffer a large drop in real applications.
 For instance, a well trained deep face model from
CASIA \cite{yi2014learning} which achieves 97.3\% recognition rate on LFW \cite{huang2007labeled} only gets 85.9\% recognition rate on our testing set of
Asian faces.
An ideal approach to this problem is to train or finetune the deep model with enough labeled data of the target domain.
However, labeling such a dataset manually is very tedious and costly.

On the other hand, with the rapid development
of the Internet, there are numerous weakly labeled data of the target domain readily available on the
Internet.
It will be quite attractive if we can obtain a cleaned dataset from this weakly labeled dataset even at a relative low recall rate, say 0.5 as the size of the original dataset is considerably large.

Note that previous works also applied some methods to
automatically clean the weakly dataset before final manual processing. For example, Dong Yi
initialized a set with some seed images and then expand this set by selecting face images verified as from the
same identity of the seed images by a pretrained face model \cite{yi2014learning}.
The main issue is a pretrained model usually does not fit the target domain well and the recall rate will be very low if we want to ensure the purity. 

We observe that in weakly labeled datasets, an identity often contains tens of face images which provides an
nice property of continuity, i.e. one face example often has a close enough neighbor with small
variance, which can be reliably found by an existing face model even though this model is trained from a
different domain. Once such links are established, we can traverse the subgraph to get a cleaned face
set for each identity. Then using the cleaned dataset, we can obtain
a face recognition model which fits the target domain better. This new model will further give us a larger cleaned dataset if we run the filtering process again. 

The key ingredients  that make our method differ from the previous works are: 1) we use the subgraph as the cleaned dataset for one identity rather than only collect one-hop neighbors of the seed samples; 2) we run the filtering process in recursive manner to gradually expand the cleaned dataset which is reasonable as the updated model fits the target domain better. 

One concern of our method is that the cleaned dataset might lack of variance as the linked faces are very close in appearance to ensure the quality. Fortunately, the variance accumulates when we traverse the graph along a path which means the cleaned dataset still contain face images of large variances.

We collect 530,560 face images from the Internet using 7,962 Asian celebrity names as queries. By running our method on this dataset, we get a cleaned dataset (purity 99.7+\%) of size 223,767.  This final cleaned dataset gives us a new face model which achieves recognition rate 93.1\% on our Asian testing dataset where the initial model trained by CASIA only achieves 85.9\%. To the best of our knowledge, our dataset is the largest dataset particularly designed for Asian face recognition task. We will publish our datasets including the original and cleaned ones soon.

In summary, our contributions are mainly two folded.
\begin{itemize}
	\item We  propose a novel method to automatically and incrementally build face datasets from a weakly labeled dataset with the help of
	an existing face recognition model.
	\item We provide large Asian face datasets designed for the study of domain specific face recognition problem.
\end{itemize}

The remaining part of this paper is organized as follows. In section II, we review the related work.
In section III, we describe how to build a cleaned dataset from the weakly labeled dataset iteratively.
In section IV, we describe our weakly labeled dataset crawled from the Internet.
In section V, we describe the face recognition model applied by our method and give the detailed network architecture. 
Section VI,
we demonstrate the effectiveness of our method by several experiments.

\section{Related Work}

The related work to our method can be roughly divided into three groups as follows.

\subsection{Face Datasets}
Face datasets play a critical role for face recognition. In the early years, face datasets are relatively small
and obtained in controlled environments, e.g. PIE \cite{sim2002cmu}, FERRET \cite{phillips2000feret}
which are designed to study the effect of particular parameters. In order to reflect the real-world challenges
of face recognition, Huang built a dataset named LFW, i.e. labeled face in the wild
\cite{huang2007labeled}, which contains 13,233 images with 5749 subjects, collected from the Internet with large
variance in pose, light and view condition. This dataset has greatly advanced the progress of face community.
Using the name list of LFW, Wolf et al. constructed a larger dataset, called YTF \cite{wolf2011face}
from the videos of YouTube. As the videos are highly compressed, YTF provides an image set of lower
quality for performance evaluation. In order to study the problem of face recognition across ages, researchers also constructed
a dataset, called CACD \cite{chen2014cross}. It includes 163,446 images of 2,000
subjects. However, only a small part of this dataset was manually checked.

Recently, with the success of deep models, the community has begun to use large scale datasets to train their networks.
Typical datasets include CelebFace of CUHK \cite{DBLP:conf/iccv/SunWT13}, SFC of Facebook \cite{taigman2014deepface} and
WDRef of Microsoft \cite{chen2012bayesian}. However, these datasets are all not public, which makes the
fairly comparison of different models very difficult.

In order to fill this gap, a large scale public dataset, CASIA \cite{yi2014learning} was
provided by Dong et al. This dataset contains 500,000 images of 10,000 celebrities collected from
IMDb website. Similarly, an even larger dataset, called MS-Celeb-1M has been proposed to
advance the community \cite{DBLP:journals/corr/GuoZHHG16}. A common property of these datasets
is that the face images are usually from western celebrities and models trained by these datasets are less optimal on
eastern faces.


\subsection{Face Recognition Models}
Compared to datasets, face recognition has gained much more attention from shallow models to deep models. The shallow
models, e.g. Eigen Face \cite{turk1991eigenfaces}, Fisher Face \cite{belhumeur1997eigenfaces},
Gabor based LDA \cite{liu2002gabor} and LBP based LDA \cite{li2007illumination} usually rely on
raw pixels or hand-crafted features and are evaluated on early datasets in controlled environments. Recently, a
set of deep face models have been proposed and greatly advanced the progress
\cite{taigman2014deepface,sun2014deep,yi2014learning,DBLP:FaceNet}. Deep face \cite{taigman2014deepface} applies 3D
alignment to warp faces to frontal views and learn deep face representations with 4,000 subjects. DeepID
\cite{sun2014deep} uses a set of small networks with each network observing a patch of the face
region for recognition. FaceNet \cite{DBLP:FaceNet} is another deep face model proposed recently, which are
trained by relative distance constraints with one large network. Using a huge dataset, FaceNet achieves
99.6\% recognition rate on LFW.

\subsection{Transfer Learning}
Transfer learning has been long studied due to its importance in practice \cite{quattoni2008transfer}.
Recently, several approaches have also been proposed for face verification. Xudong proposed to use Joint Bayesian model with
KL regularization where only a limited number of training examples of target domain are available \cite{cao2013practical}.  Xiaogang
et al. proposed an information-theoretic approach to narrow the representation gap between photos and sketches \cite{zhang2011coupled}.
Though the direct output of our method is a dataset, we can obtain a new model immediately by applying
this dataset to train or finetune a model, which serves the same goal as transfer learning.


\section{Method Overview}

\begin{figure*}[!htb]
	\begin{center}
		\includegraphics [width=5.6 in]{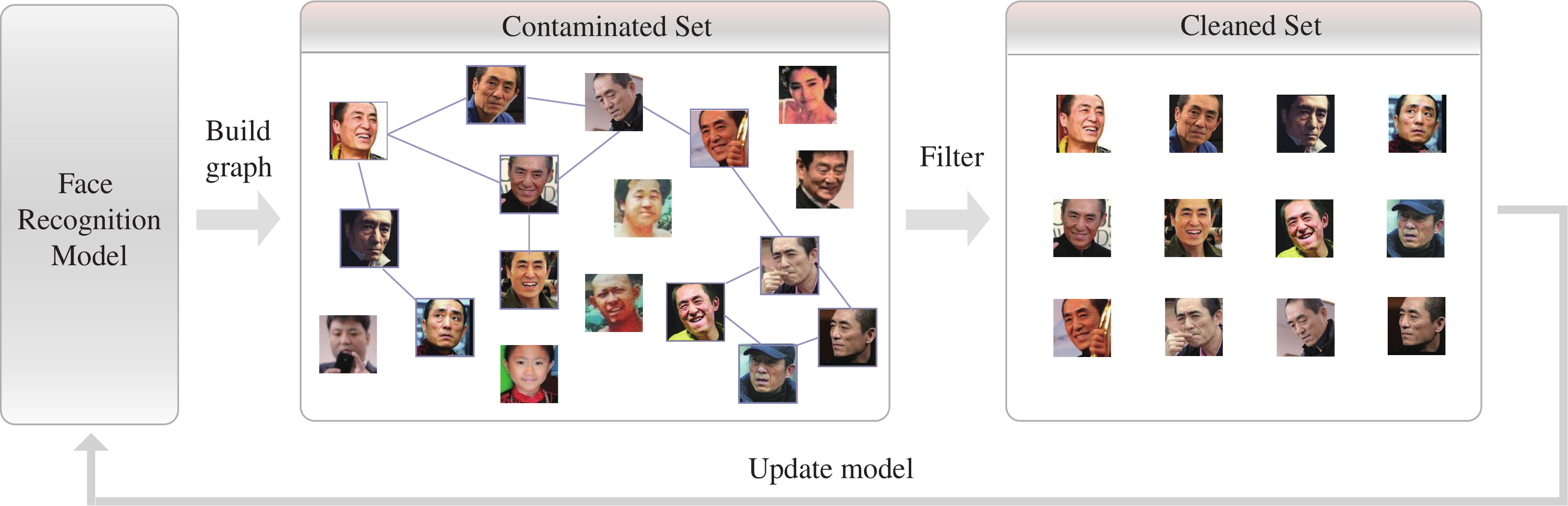}
		\caption{Illustration of data cleaning method. Face images in contaminated set linked by an edge
			are verified as from the same identity by the existing model. The maximal subgraph is
			collected as the cleaned dataset for the identity which is further used to update the model. 
		}
		\label{fig:ModelMap}
	\end{center}
\end{figure*}

In this section, we describe the overall principle of our method. First, we give a formal description of our problem. We are given an existing face model, denoted
by its parameter set $W_s$ trained by a dataset $D_s$, which can
produce similarity or distance score $d(I_i,I_j;W_s)$
for face images $I_i$ and $I_j$. In addition, we are given
a large amount of weakly labeled dataset $D_t$, which are mainly drawn from a different
domain, e.g. a different race. By weakly, we mean the majority of the images are
correctly labeled while a small portion are wrong. The direct goal of our method is
to build a clean dataset $D_t'$ from $D_t$ which is to serve the ultimate goal
of getting a new model $W_t$ for this new domain.

Our method is based on the continuity structure of face images of one identity. That
is, given a face image of one identity, we can often find a close enough
neighbor image of the same identity  in the weakly labeled dataset. As such neighbors
have small variance to the query one, they can be easily found by an existing
face model with high confidence, even though this model is trained from a different domain.
In terms of graph representation, we can create a graph for each identity with the
edges linking the matched face images under the help of an existing face recognition model
using a tight threshold $T$. Once such graph has been constructed, we can collect a
maximal subgraph as the cleaned dataset for that identity. 
The reason why we use the
maximal subgraph is we assume the majority of the images associated with the label are from the same identity. Obviously, the quality of the cleaned dataset $D_t'$ highly
depends on the value of the threshold $T$. Large $T$ leads to a high recall
of the correct face images while increasing the risk of introducing wrong images, and small
$T$ ensures the purity of the dataset $D_t'$ at the price of missing more correct samples.

As the face model plays a central role in constructing the graph, a model finetuned on the cleaned set is supposed to give better filtering result. Thus we can repeat the filtering process to incrementally obtain a large scale cleaned dataset. Figure \ref{fig:ModelMap} shows the overall principle.

If we assume that the face image which has
the most neighbors is correctly labeled (in most cases, this assumption holds), then the cleaning process can be simply implemented by Algorithm
\ref{alg:data-cleaning}.

\begin{small}
\begin{algorithm}[htb]
	\caption{Data cleaning process for one identity}
	\label{alg:data-cleaning}

	\SetKwInOut{Input}{Input}
	\SetKwInOut{Output}{Output}
	\Input{Contaminated face set $G$, a trained deep face model $M$}
	\Output{Clean dataset $S$ for the identity}

	Create a selected set $S$ and a remaining set $R$\;
	Find an anchor face $I_0$ which has the most neighbors\;
	Add $I_0$ to $S$ and set $R=G-S$\;
	\For{$I \in R$} {
		\For{$J \in S$} {
			\If{$M$.match($I$,$J$)=TRUE} {
				Add $I$ to $S$ and remove $I$ from $R$\;
				Break\;
			}
		}
	}
\end{algorithm}
\end{small}

One concern of our method is the cleaned dataset might lack of enough variance as
the linked faces are close in appearance. Fortunately, as seen from Figure \ref{fig:ModelMap}, the variance
can accumulate along the path of the graph. Once a training sample (triplet wise or pair wise constraint) contains two faces
connected by several hops, then the intra-class variance of such sample is still considerable. The exact training form of face recognition models will be discussed later.

\section{Weakly Labeled Data Collection}

In this section, we describe the weakly dataset crawled from the Internet. As we focus on Asian faces, we use Baidu, the biggest search engine in China
to search images. More specifically, our data collection process has two steps. First, we obtain
a name list of Asian celebrities from the the search engine which is automatically provided
when searching an Asian celebrity. Then for each name in the list, we query the
search engine and use the top $N$ images as the weakly labeled data. The number
$N$ usually ranges from 30 to 100 as the crawl process is not stable, which
is caused by expiration of the target or unreachability of the network.
Figure \ref{fig:ModelMap} shows some typical examples of one identity. We summarize the data characteristics as follows.

\textbf{Quality} The quality of most images are relatively high, i.e. high resolution (more than
1M byte) and  good sharpness. 

\textbf{Purity} For famous celebrities, about 85 percent of the images are correctly associated with the
query name in the top 100 images.

\textbf{Variance} The variances of faces caused by different pose and light conditions are obvious as
we can see from the samples. Usually the yaw, pitch and roll angles range from -15 to 15.

\textbf{Continuity} Most of the face images usually have a close neighbor, i.e. another face image of the same identity that has small variance. This is critical for our method as we want to build a connected subgraph for each identity with a tight threshold.

For all the face images, we use a face detector implemented by ourselves base on
deep CNN models to crop the faces.
After this step, we finally get a large dataset containing 530,560 face images from 7,962 identities.
We call this dataset CACFD (Contaminated Asian Celeb Face
Dataset). This dataset is further cleaned by the aforementioned method which will be discussed in
the experiments.

\section{Face Recognition Model and Architecture}
\subsection{Face Recognition Model}
In this section, we make an introduction to the deep face recognition model adopted in
our method, i.e. triplet based feature embedding model \cite{DBLP:FaceNet,ding2015deep}. Actually, the way we designed
to purify the weakly labeled datasets also holds for other face recognition models such as
pair based models.  In triplet based recognition model, the network is trained by a
set of relative distance constraints organized by triplets. Each triplet contains three images denoted as
$O_i^1,O_i^2,O_i^3$, with $O_i^1$ and $O_i^2$ from one subject and $O_i^3$ from another subject. We use
$W$ to denote the network parameter set and $F_W(I)$ to denote the output feature for
image $I$ produced by the network. Essentially, triplet based face model is to solve the
network parameter set $W$ to satisfy the following distance constraints, i.e. distance between matched faces
should be smaller than the distance between mismatched faces:
\begin{equation}
||F_W(O_i^1)-F_W(O_i^2)||<||F_W(O_i^1)-F_W(O_i^3)||
\end{equation}

This constraints are further turned into a hinge-loss like objective function $f$ where $C$ is a margin value and $O=\{O_i\}$ is the triplet set. This objective can be solved
efficiently using image-based gradient descent algorithm \cite{ding2015deep}.

\begin{equation}
\begin{split}
f(W,O)=\Sigma_{i=1}^{n}\max\{& ||F_W(O_i^1)-F_W(O_i^2)||^2 \\
& - ||F_W(O_i^1)-F_W(O_i^3)||^2, C\}
\end{split}
\label{eq:triplet_loss}
\end{equation}

\begin{figure}[!htb]
	\begin{center}
		\includegraphics [width=2.8 in]{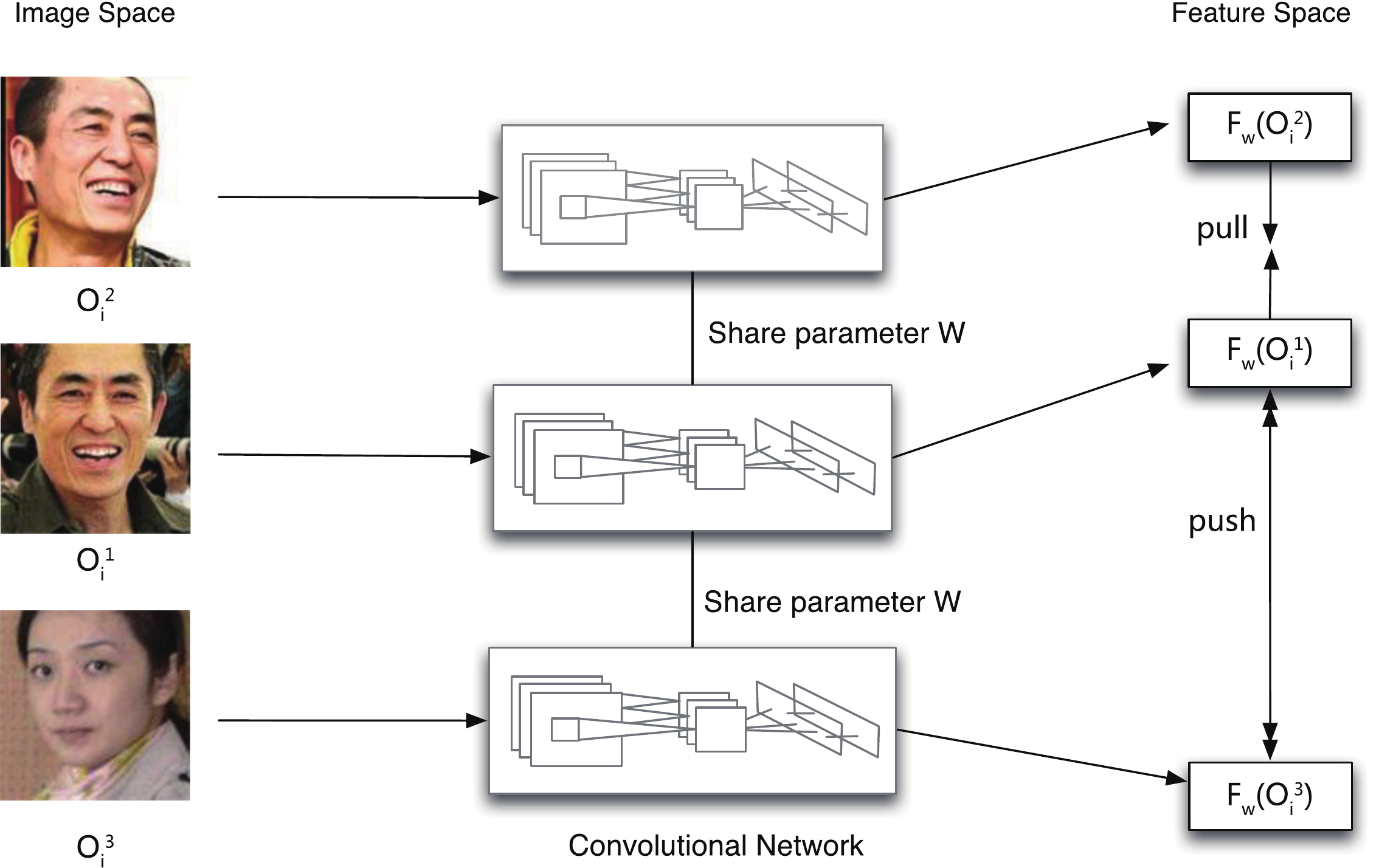}
		\caption{Illustration of representation learning using triplet constraints. This model requires the Euclidean distance between matched
			faces is smaller than the distance between mismatched faces in the feature space.}
		\label{fig:triplet-model}
	\end{center}
\end{figure}

From the definition of the objective, the triplets play a critical role for the model
performance. Usually, triplets are generated from labeled datasets. Given a labeled dataset, theoretically we can
enumerate all the triplets according to the definition.
However, it is impossible to use all the triplets to train the model due to
the exponentially growing number of triplets and limited memory. Instead, we still need to apply
SGD algorithm to solve the parameters iteratively, i.e. select a batch of triplets and update
the parameter with the gradient derived from the batch.
There are several means to construct the batch of the triplets in each iteration. For
instance, for each triplet, we can randomly select $O_1$, $O_2$ and $O_3$ according to the
definition. Note for a large labeled dataset, the number of distinct images in the triplets
are about three times the size of triplets as the probability of different triplets sharing
the same image is low. In other words, only a few distance constraints are applied
on the selected images in each batch, we call this sparse triplet generation policy. In
contrast to this sparse policy, we can first select a small number of identities with
each identity using a fixed number of face images and enumerate all the possible triplets
from the selected images. We call this dense sampling policy.
As proved in Ding's work, there exists an algorithm in which the computational cost mainly depends
on the size of the distinct images in triplets. Thus the dense sampling policy has a
remarkable advantage over the sparse policy as all the possible distance constraints are applied to
the selected images \cite{ding2015deep}.

\subsection{Network Architecture}
\begin{figure*}[htb]
	\begin{center}
		\includegraphics [width=5.6 in]{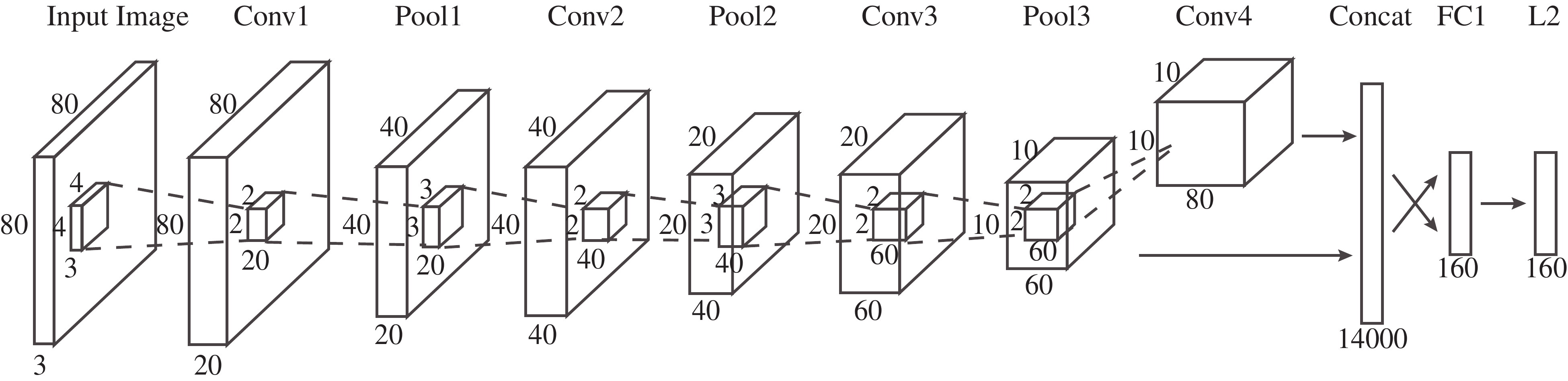}
		\caption{Architecture of one network. }
		\label{fig:network_arch}
	\end{center}
\end{figure*}

In this section, we describe the network architecture which is used by our triplet model.

\begin{figure}[htb]
	\begin{center}
		\includegraphics [width=3.2 in]{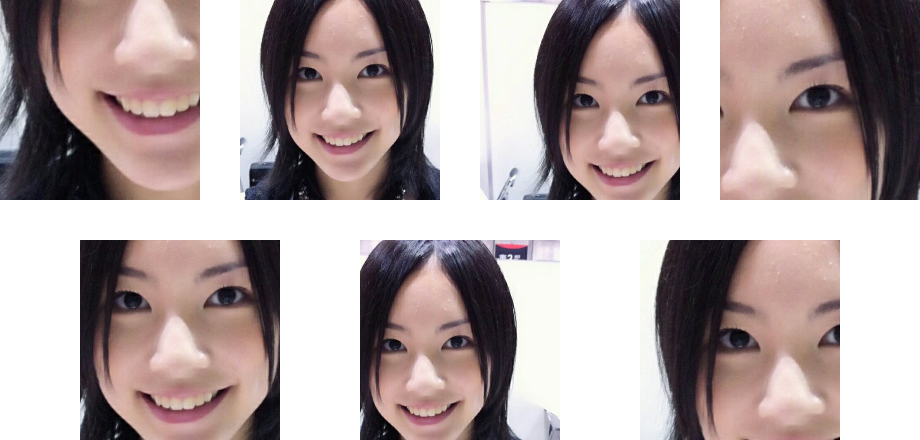}
		\caption{Illustration of different patches used by our ensemble model.}
		\label{fig:model_patches}
	\end{center}
\end{figure}

We use multiple small networks to obtain our final feature with each network taking a
particular patch of a face image as input as in DeepID \cite{sun2014deep} rather than
a large network.
Each network is trained by the triplet loss objective as in Equation \ref{eq:triplet_loss}. We
argue that this ensemble approach has several advantages. 1) A small network can be trained
much faster than a huge network. Thus the ensemble model can be easily trained in
parallel when multiple GPUs are available;  2) Inputs can be better aligned as the
selected patches are usually centered at the facial keypoints. Based on this ensemble model,
we use 7 square patches of size $80 \times 80$ with each patch corresponding to a particular scale and location which are shown in Figure
\ref{fig:model_patches}. The 7 networks share the same architecture 
as in Figure \ref{fig:network_arch}. In this architecture, there are 10 layers including the final $L2$ normalization layer which is to restrict the feature on a unit sphere. We give the detailed parameter configurations in table \ref{table:ensemblenet_config} . During the testing stage, given a face image, we get a set of patches and feed these patches to the corresponding networks to obtain a concatenated feature
($160 \times 7 = 1120$ dimensional). Then we apply PCA to get a 300 dimensional feature as the final
feature for this face image.

\begin{table}
	\centering
	\begin{tabular}{|c|c|c|c|c|}
		\hline
		\thead{ Layer} & \thead{Input\\size} & \thead{Output \\ Size} & \thead{ Kernel size, \\ Stride } \\
		\hline
		conv1 & $80\times80\times3$  & $80\times80\times20$ & $4\times4\times3,1$  \\
		\hline
		pool1 & $80\times80\times20$ & $40\times40\times20$ & $2\times2\times20,2$ \\
		\hline
		conv2 & $40\times40\times20$ & $40\times40\times40$ & $3\times3\times20,1$ \\
		\hline
		pool2 & $40\times40\times40$ & $20\times20\times40$ & $2\times2\times40,2$ \\
		\hline
		conv3 & $20\times20\times40$ & $20\times20\times60$ & $3\times3\times40,1$ \\
		\hline
		pool3 & $20\times20\times60$ & $10\times10\times60$ & $2\times2\times60,2$ \\
		\hline
		conv4 & $10\times10\times60$ & $10\times10\times80$ & $2\times2\times60,1$ \\
		\hline
		concat & $14000$ & $14000$ &\\
		\hline
		fc1 & $14000$ & $160$ & \\
		\hline
		L2 & $160$ & $160$ & \\
		\hline
	\end{tabular}
	\caption{Network configurations of input $80 \times 80$.}
	\label{table:ensemblenet_config}
\end{table}

\section{Experiment}
In this section, we evaluate the effectiveness of our method from two aspects, i.e. the
data purity and the recognition performance of the models trained by the cleaned dataset. 
\subsection{Pretrained Deep Face Models}

We use CASIA to train a deep face model using the aforementioned network architecture as
the pretrained recognition model. More specifically,
we train 7 networks using the architecture specified in
table \ref{table:ensemblenet_config} with each network observing a different patch as depicted in Figure \ref{fig:model_patches}. We
use dense sampling scheme to generate the triplets and solve the parameters with image based
fast SGD algorithm. In each iteration,
we select 10 subjects with each subject using 30 images, i.e. 300 distinct images per batch in total.
We stop training after 500,000 iterations when the training process basically converges, which takes about two days on a server equipped with GRID K520 GPUs. We combine the
features of different networks and use PCA to reduce the dimensionality to 300. The 
recognition rate on LFW testing set is 97.3\%.

\subsection{Evaluation by Purity}
As we mentioned, the purity of the processed data depends on the threshold of the
governing face model. Better purity comes at the price of less coverage with a tight
threshold. Actually, this characteristic can be quantitively measured by the widely adopted PR (precision-vs-recall) curve, i.e.
precision/purity vs recall. More precisely, given a weakly labeled dataset and a matching threshold, we can get a filtered image set. If we know the ground-truth label of each image,
then we can find out how many images are correctly labeled in the filtered set
and how many correct images are missed from the set. With $D_t$ to
denote the weakly labeled dataset and $D_t'$ to denote the cleaned dataset,
then we can define precision and recall as follows:
\begin{equation}
\mathrm{precision}=\frac{ |\{\mathrm{correctly\ labeled\  faces \ in\ } D_t'\}| }{ | \{ \mathrm{ faces\ in\ } D_t'\} | }
\end{equation}
\begin{equation}
\mathrm{recall}=\frac{ |\{\mathrm{correctly\ labeled\  faces \ in\ } D_t'\}| }{ | \{\mathrm{correctly\ labeled\  faces \ in\ } D_t\} | }
\end{equation}

As it is very labor intensive to label all the face images, we randomly select
325 subjects (20108 face images in total) and manually label the images of these subjects for statistics.

\begin{figure}[htb]
	\begin{center}
		\includegraphics [width=3.2 in]{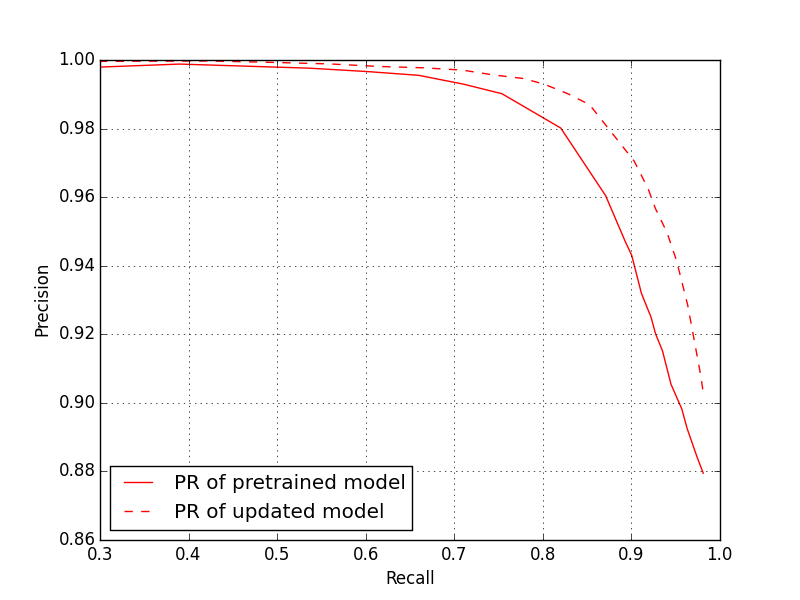}
		\caption{PR curves of the filtering process of two models.}
		\label{fig:pvsr-1}
	\end{center}
\end{figure}

We run the cleaning process with two iterations. The first iteration filters the dataset with a pretrained model from CASIA. The second iteration filters the dataset with an updated model trained by the filtered dataset (purity 99.8\% and recall 53.6\%) of the first iteration. We give the corresponding PR curves in Figure \ref{fig:pvsr-1}, in which the solid line corresponds to the filtering process of CASIA model and the dashed line corresponds to  the filtering process of the updated model respectively. As we expected, the second filtering process gives a  higher recall of 61.5\% at precision 99.8\%. This clearly demonstrates the advantages of our iterative filtering method over the previous works which only filter the dataset once. Table  \ref{table:pvsr} lists precision at different recall rates.
\begin{table}[tpb]
	\centering
	\begin{tabular}{|c|c|c|}
		\hline
		\thead {Precision} & \thead {Recall of \\ pretrained model} & \thead {Recall of \\ updated model} \\
		\hline
		99.9\% & 39.0\% & 55.9\% \\
		\hline
		99.8\% & 53.6\% & 61.5\% \\
		\hline
		99.7\% & 60.1\% & 70.9\% \\
		\hline
		99.6\% & 66.0\% & 74.4\% \\
		\hline
		99.0\% & 75.4\% & 82.8\% \\
		\hline
		95.0\% & 88.8\% & 94.0\% \\
		\hline
		90.2\% & 95.0\% & 98.2\% \\
		\hline
	\end{tabular}
	\caption{Precision vs recall of pretrained model and updated model.}
	\label{table:pvsr}
\end{table}

\subsection{Data Evaluation by Recognition Performance}
\begin{figure}[htb]
	\begin{center}
		\includegraphics [width=3.2 in]{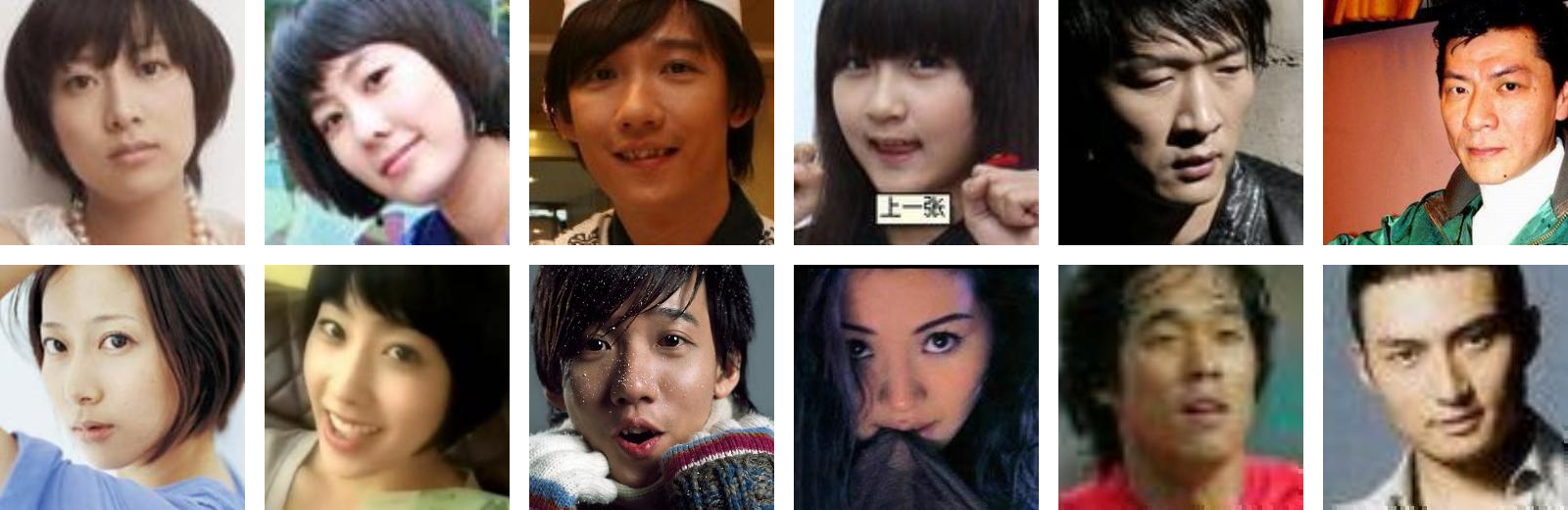}
		\caption{Some testing pair samples. The left three columns are positive pairs and the remainings are negative pairs.}
		\label{fig:test_pairs}
	\end{center}
\end{figure}

The underlying goal of creating a dataset is to obtain new models for the target
domain. In this part, we evaluate how our cleaned dataset benefits the face model for the target domain. Thus we first create a benchmark testing set of the target domain and adopt the similar evaluation
protocol as LFW \cite{huang2007labeled}.  More specifically, we use the same 325 subjects manually labeled for purity evaluation which are removed from the training data to construct the testing dataset. 
We construct 25,000 positive pairs and 25,000 negative pairs for final performance report. Figure \ref{fig:test_pairs} lists some typical testing pair examples with each column representing one pair (left three columns are positive pairs and remainings are negative ones). We can see that the testing pairs are quite hard even for human to verify whether they are from the same identity.

We use the same architecture of the pretrained model for evaluation models. We first optimize the model parameters with the second-round cleaned dataset of size 223,767 (recall 70.9\%) and purity 99.7\%. We follow almost the same learning strategy as for the pretrained model, i.e. we select a batch of face images and use dense sampling policy to generate triplets. We stop the learning process after 200,000 iterations. We concatenate the features and run PCA to reduce the dimensionality to 300. Using this 300 dimensional feature, the recognition rate of our new model reaches 93.1\% on our testing set where the model pretrained by CASIA only achieves 85.9\%. 

As a comparison, we also optimize the parameters of the model using the first-round cleaned dataset of size 160,875 at recall 53.6\%. After 200,000 training iterations, we get a recognition rate of 92.8\% on our testing set. Table \ref{table:error_rate} lists accuracies of models
trained by different datasets, which clearly shows the effectiveness of our method.

\begin{table}[tpb]
	\centering
	\begin{tabular}{|c|c|c|}
		\hline
		\thead {Training\\Dataset} & \thead {Size} & \thead {Recognition rate} \\
		\hline
		CASIA & 500,000 & 85.9\% \\
		\hline
		\shortstack{Cleaned set by \\ pretrained model} & 160,875 & 92.8\% \\
		\hline
		\shortstack{Cleaned set by \\ finetuned model} & 223,767 & 93.1\% \\
		\hline
	\end{tabular}
	\caption{Face recognition rates of models trained with different datasets.}
	\label{table:error_rate}
\end{table}

\section{Conclusion}
In this paper, we propose a novel method to automatically build clean face datasets from
a weakly labeled dataset of a new domain. We iteratively filter the original dataset by a model trained with the cleaned dataset in last iteration.
By starting from a deep face model trained by CASIA, we get an almost cleaned dataset of size 223,767 from
530,560 face images of 7,962 Asian celebrities after two iterations. Using this cleaned dataset,
we get a face model whose recognition rate reaches 93.1\% on the testing set of Asian faces where the pretrained model only achieves 85.9\%.

\appendix

\bibliographystyle{aaai}
\bibliography{references}

\end{document}